%% file: main.tex
\documentclass[11pt,twocolumn,letterpaper]{article}

\usepackage[left=0.8in,right=0.8in,top=0.8in,bottom=0.8in]{geometry}
\usepackage{cite}
\usepackage{amsmath,amssymb,amsfonts}
\usepackage{algorithmic}
\usepackage{graphicx}
\usepackage{textcomp}
\usepackage{xcolor}
\def\BibTeX{{\rm B\kern-.05em{\sc i\kern-.025em b}\kern-.08em
    T\kern-.1667em\lower.7ex\hbox{E}\kern-.125emX}}
\usepackage{booktabs}
\usepackage{graphicx}
\usepackage{amsmath}
\usepackage{amssymb}
\usepackage{booktabs}
\usepackage{blindtext}
\usepackage{hyperref}
\usepackage{comment}
\usepackage{subcaption}
\usepackage{xspace}
\usepackage{bm}
\usepackage{amsfonts}
\usepackage{graphicx}
\usepackage{multirow}

\usepackage[capitalize]{cleveref}
\crefname{section}{Sec.}{Secs.}
\Crefname{section}{Section}{Sections}
\Crefname{table}{Table}{Tables}
\crefname{table}{Tab.}{Tabs.}

\makeatletter
\DeclareRobustCommand\onedot{\futurelet\@let@token\@onedot}
\def\@onedot{\ifx\@let@token.\else.\null\fi\xspace}

\def\eg{\emph{e.g}\onedot} 
\def\ie{\emph{i.e}\onedot}

\def\etal{\emph{et al}\onedot}
\makeatother

\definecolor{blue}{RGB}{0,50,200}
\definecolor{contrast-blue}{RGB}{52, 89, 237}
\definecolor{contrast-orange}{RGB}{230, 126, 0}
\definecolor{contrast-purple}{RGB}{121, 0, 201}

\begin{document}

\title{SAGE: Saliency-Guided Contrastive Embeddings}

\author{
  Colton R. Crum \\
  University of Notre Dame \\
  \tt\small{ccrum@nd.edu}
  \and
  Christopher Sweet \\
  University of Notre Dame \\
  \tt\small{csweet1@nd.edu}
  \and
  Adam Czajka \\
  University of Notre Dame \\
  \tt\small{aczajka@nd.edu}}

\date{}

\maketitle

\input{latex/00-Abstract}
\input{latex/01-Introduction}
\input{latex/02-Related-Work}
\input{latex/03-Methodology}

\input{latex/04-Results}

\input{latex/05-Conclusion}

{\small
\bibliographystyle{ieee}
\bibliography{main}
}

\end{document}

%% file: latex/00-Abstract.tex
\begin{abstract}
Integrating human perceptual priors into the training of neural networks has been shown to raise model generalization, serve as an effective regularizer, and align models with human expertise for applications in high-risk domains. Existing approaches to integrate saliency into model training often rely on internal model mechanisms, which recent research suggests may be unreliable. Our insight is that many challenges associated with saliency-guided training stem from the placement of the guidance approaches solely within the image space. Instead, we move away from the image space, use the model's latent space embeddings to steer human guidance during training, and we propose SAGE (Saliency-Guided Contrastive Embeddings): a loss function that integrates human saliency into network training using contrastive embeddings. We apply salient-preserving and saliency-degrading signal augmentations to the input and capture the changes in embeddings and model logits. We guide the model towards salient features and away from non-salient features using a contrastive triplet loss. Additionally, we perform a sanity check on the logit distributions to ensure that the model outputs match the saliency-based augmentations. We demonstrate a boost in classification performance across both open- and closed-set scenarios against SOTA saliency-based methods, showing SAGE's effectiveness across various backbones, and include experiments to suggest its wide generalization across tasks.
\end{abstract}

%% file: latex/01-Introduction.tex
\section{Introduction}
\label{sec:sage-introduction}
\input{figures/method-example}

\subsection{Background}
Saliency-based training incorporates saliency maps, often derived from human experts, into the training of neural networks \cite{boyd2021cyborg, crum2023mentor, linsley2018learning}. Methods focus on guiding models to salient regions of the input image, away from spurious features that can be correlated with class labels, but otherwise erroneous in solving the overarching task. Saliency maps generated by the model's internal mechanisms are compared with the human saliency map, where the model is penalized for its deviation from the human saliency map. Incorporating such human oversight into model training is often necessary and is motivated by the severity of mission critical tasks or to challenges associated with model training. Examples of the former include high-risk domains where humans must be within the loop. The former includes open-set scenarios, highly imbalanced training data, or a limited number of training samples. Thus, saliency-based techniques are often found within the context of medicine, biometrics, robotics, and autonomous vehicles.

\subsection{Motivation}
These techniques, however, are constrained by their ability to convey information accurately, as this forms the underlying representation that guides the network during training. Thus, saliency maps form an upper bound on the potential for saliency-based methods. Mounting research suggests that model saliency may be unreliable, where saliency maps are not always reflective of the model's classification mechanism \cite{adebayo2018sanity}. If the model's internal saliency map is compromised, any effort to steer the model towards salient features is severely undermined, or potentially outright misleading. Others have demonstrated that the means of generating model saliency (\eg activations, gradients, etc.) can be easily manipulated, which implies that these attacks thwart the use of saliency in model training. Furthermore, saliency maps are limited in their ability to convey model outputs due to their low resolution. Their reliance on internal mechanisms constrains the saliency map's resolution to a mere fraction of the input resolution (\eg 3\% of the input). Although this may be managed from rescaling or projections \cite{draelos2020hirescam}, it does so by improving visualization quality, not in the quantity of information portrayed. Together, these challenges cripple the validity of saliency-guided training.

\subsection{Proposed Approach}
Instead of forcing the model to align explanations in image space using \eg, $L_2$ norm, we avoid the above pitfalls by guiding the model towards salient features in the model's own embedding norm. Guidance within the model's embedding space preserves the richness of high-dimensional information, enables saliency at full resolution, and supports a broad range of network architectures and applications. This paper presents \textbf{SAGE}, \textbf{SA}lient-\textbf{G}uided contrastive \textbf{E}mbeddings, which applies saliency-based augmentations that measure changes within the model's embeddings and logits. These changes delineate an underlying structure that can be used to steer embeddings with a contrastive triplet loss and to ensure that model logits reflect these changes by penalizing the divergence between the respective augmentations. The proposed method guides the model towards human-salient features for any neural network architecture using a single, differentiable loss function, without requiring architectural modifications. We achieve state-of-the-art performance across several human-salient datasets, saliency-based methods, and network backbones. Finally, we demonstrate how our method generalizes across a wide range of applications and how our method can be adapted and implemented at scale.

%% file: figures/method-example.tex
    \begin{figure*}[t]
      \centering
      \includegraphics[width=0.95\linewidth]{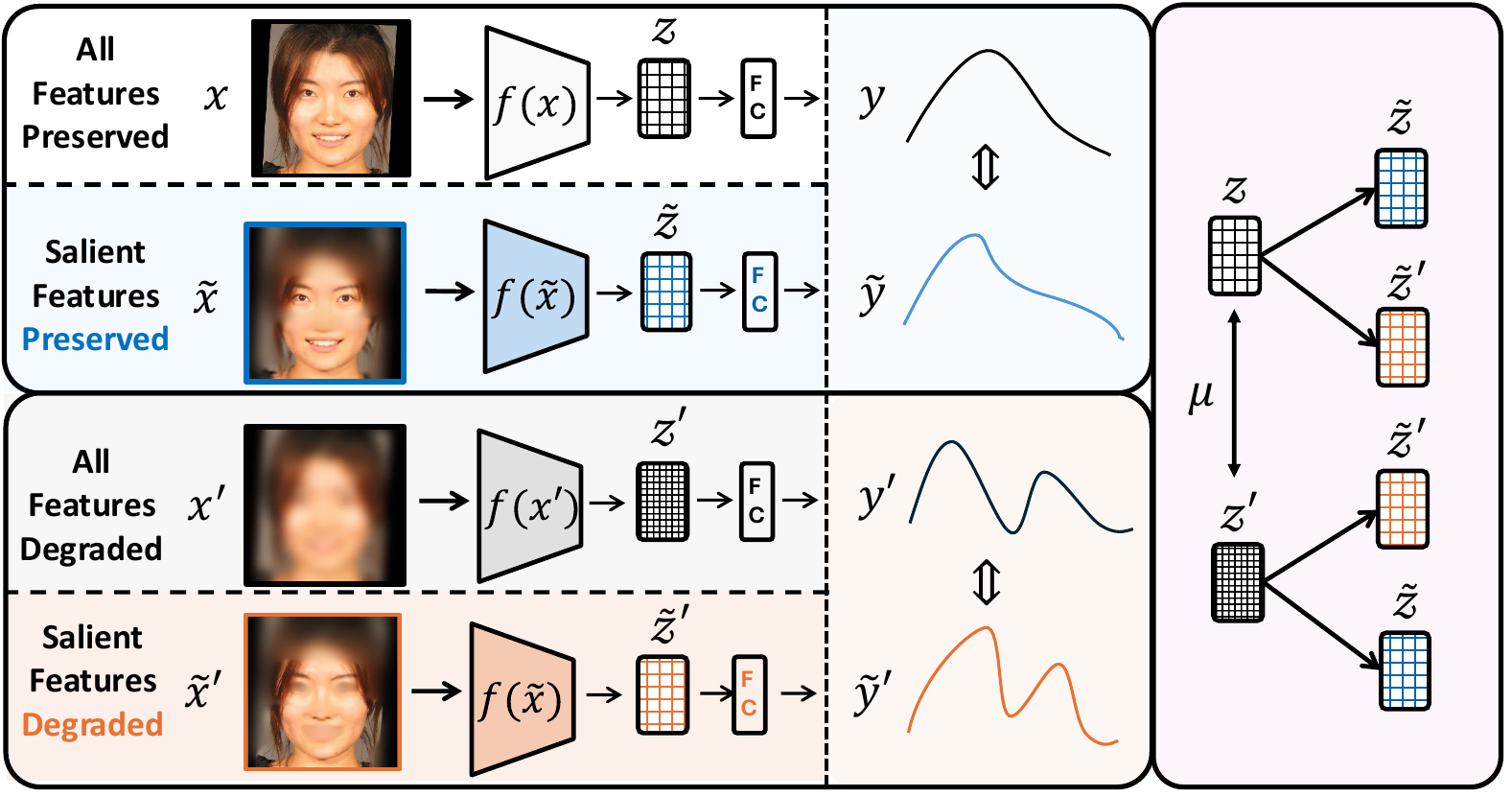}
      
      \caption{An overview of the two main components of \textbf{SAGE}. Saliency-based Augmentations (left): Input $x$ is blurred to generate $x'$, a degraded image that removes high-frequency information. Using salient features indicated by human experts, we apply selective blurring to generate $\tilde{x}$ to \textcolor{contrast-blue}{\bf preserve salient features}, and $\tilde{x}'$ to \textcolor{contrast-orange}{\bf degrade salient features}. During training, the model makes a forward pass for each $x$ with $y$ output logits and $z$ embeddings captured by a predefined hook. Our loss encourages an alignment of logit distributions (middle column) and guides respective \textcolor{contrast-purple}{\bf embeddings} using a contrastive triplet loss (right).}
      \label{fig:sage-overview}
    \end{figure*}

%% file: latex/02-Related-Work.tex
\section{Related Work}
\label{sec:sage-related-work}

\subsection{Interpretable and Explainable AI}

Since the conception of neural networks, researchers have sought to understand model decision-making, as a highly competent classification model is futile in high-risk tasks if human operators cannot interpret its decisions. As a result, explainable, interpretable, and trustworthy AI has spawned a copious of research methods devised to visually indicate ``where the model was looking'' when it arrived at a certain classification decision. Within classification models, two main approaches have been explored to extract pertinent information from the model: (a) ``white-box'' and (b) ``black-box'' approaches. ``White-box'' techniques utilize the model's internal mechanisms such as its feature map activations \cite{zhou2016learning, wang2020score, wang2019score, wang2020ss} or gradients \cite{selvaraju2017grad, chattopadhay2018grad} to produce a model saliency map. Other methods rely on principle component analysis, image-space perturbations, or second-order gradients to formulate more fine-grained visuals. Investigations into the fidelity and trustworthiness of saliency maps have led to questions of their true reflection of model intuition, leading many researchers to either rectify these observations \cite{fu2020axiom}, or turn towards an entirely different paradigm of model salience altogether. ``Black-box'' approaches look beyond the internal mechanisms and instead observe the system as a whole. These methods opt out of utilizing the model's internal mechanisms (\eg activations, gradients), and instead build their saliency maps by modifying the input and measuring the differences within the output \cite{petsiuk2018rise}.

RISE generates model saliency by selectively perturbing regions of the input and recording changes within the differences within the softmax scores. Unlike nearly all saliency methods which focus on a localized region of the network or neglects model inputs, RISE records saliency through a wholistic measure that observes the entire inference pipeline from input to output. Similarly, the proposed method also considers the input-to-output pipeline, but does so without generating saliency within the $L_2$ norm. Moreover, we avoid the heavy computational overhead and account for the shape of the softmax distribution with temperature scaling.

\subsection{Human Saliency-guided Training}

Human saliency-guided training aims to integrate human-salient information provided by saliency maps into model training, which has shown to raise model generalization \cite{boyd2021cyborg, crum2023mentor}, guide the networks towards salient-features, reduce the number of training samples \cite{crum2023mentor}, create more human-interpretable outputs for human operators \cite{linsley2018learning}, and improve model convergence during training \cite{crum2023mentor}. Due to their effort to align with human experts, saliency-based methods also hold several advantages within global efforts regarding AI governance and regulatory strategies \cite{crum2023seriously}.

Early works in integrating saliency into model training utilized class activation mappings (CAM) \cite{boyd2021cyborg, zhu2022gaze}. Boyd \etal proposed CYBORG, a saliency-guided loss that generates CAMs during training penalizes the model from deviating from the human saliency map \cite{boyd2021cyborg}. GG-CAM deviates from CYBORG in that it integrates CAMs into training using a modified classification embedding \cite{zhu2022gaze}. Though far less sophisticated in its saliency generation process, it has proven crucial in saliency-based methods over more complex methods due to its ability to generate model saliency in batches on a single forward pass. Other methods may require sharper visualizations by incorporating gradients, but at the computational cost of backpropagating through the network several times. Additionally, visualization can be steered off path due to their reliance on noisy and batch-dependent gradients.

Similar to CYBORG, UNET+Gaze offers a loss-based approach \cite{karargyris2021creation}, except that model saliency is generated by a dedicated decoder rather than the classifier. Notably, the use of a dedicated decoder allows for saliency to be generated at full input resolution, unlike CYBORG \cite{boyd2021cyborg}. These advantages have also been exploited as a pretraining strategy, where embeddings are trained from the model's input to output, covering the entire model instead of a single layer. MENTOR decouples the training into two stages, where the model first learns to predict the saliency corresponding to each respective image, and second the model learns the classification task. MENTOR is motivated by the idea that models must first know where to look before knowing what they are looking at. Similarly, GazeGNN implements a graph neural network to impose structure on network embeddings.

However, these methods fail to consider two critical concepts. First, the overwhelming majority of approaches consider human saliency with the $L_2$ norm, which poses issues as previously discussed. Second, methods that consider embeddings do so without enforcing continuity between model outputs, raising questions as to how saliency is being integrated into the model's understanding of a task. In other words, if we cannot understand how network's utilize this information, then it nearly invalidates the underlying reasoning behind incorporating human saliency into training. Left unchecked, it is unclear in the precise role that model embeddings play in the model's latent space, and performance gains may be an artifact of pure regularization (\ie algorithmic regularization techniques, such as dropout).

%% file: latex/03-Methodology.tex
\section{The SAGE Approach}
\label{sec:sage-methods}
\input{figures/Face-Sampler}

This section introduces \textbf{SAGE}, a loss-based approach to saliency-guided training through network's embeddings that maintain outputs that follow human perception. Our method couples conventional wisdom from the existing literature in the xAI area \cite{crum2023explain}. First, the original model outputs and outputs for which the non-salient information is removed should be similar. Second, if the salient (important) regions of the input are removed, the corresponding outputs should be reflective of an image of low information (Sec. \ref{sec:sage-augmentations}). SAGE combines these two priors into a contrastive learning approach (Sec. \ref{sec:contrastive-embeddings}).

Furthermore, the network's internal embeddings should also follow a similar behavior. That is, in addition to build appropriate loss functions on network's outputs, analogous rules can be verified for network's internal representations (Sec. \ref{sec:logit-alignment}).

\subsection{Saliency-based Augmentations}
\label{sec:sage-augmentations}

Removing information from the input space without inadvertently creating new information is nontrivial and proven to be challenging. Convolutional neural networks have long been susceptible to latching onto sharp edges and mistaken them for features. Thus, non-salient information cannot be completely eliminated, but rather the amplitude should be reduced more smoothly over a neighborhood of the signal. Nevertheless, Gaussian blurring has shown to reduce the high-frequency information without introducing extraneous features. For other architectures, such as vision transformers, the tokenization of the input allows for simpler approaches such as pixel dimming.

Given an input image $x_i$, and an accompanying externally-derived (\eg, by a human expert) saliency map $x_h$ (indicating important features of the input), we first augment the image using a Gaussian blur so that the salient information is preserved, to create an altered image $\tilde{x}_i$. The intensity of the blurring is dependent upon the human saliency map, meaning regions of no importance received the maximum amount of blurring, whereas in salient regions no blurring is applied. Second, we augment the input image to instead degrade salient information, denoted as $\tilde{x}'$, which is generated using the inverse of the human saliency map, $x'_h$, since it indicates the non-salient regions of the input. Finally, we apply the blur to the entire input image to created a degraded image $x_i'$, to complement to the original input and serve as a way to standardize the impact blurring has on the network (see Fig. \ref{fig:sage-overview} and Fig. \ref{fig:sage-notation-dataset} for an overview). 

For our experiments, we use a fixed Gaussian blurring kernel size $K=(7,7)$, and a standard deviation $\sigma=10$, which controls the max intensity of the Gaussian blur. Other augmentations can be explored depending on the target domain and network architecture, such as adding random noise \cite{parzianello2022saliency} or dimming. We then pass these images through the network to obtain model outputs (logits) and embeddings. 

Formally, we denote these changes using the notion of a supervised classifier: $f(x_i)=y_i$ where $y_i$ is the logits vector of model $f$ given the $i$-th image in a batch. We represent network embeddings as $f_\theta(x_i)=z_i$, which are the embeddings associated with the model's parameters $\theta$ for a given input. SAGE uses model outputs to align logits to ensure the behavior is consistent with the respective salient features (Sec. \ref{sec:logit-alignment}). In addition, model embeddings are used by SAGE to steer the network towards salient features, and away from non-salient features (Sec. \ref{sec:contrastive-embeddings}).

\subsection{Saliency-based Logit Alignment}
\label{sec:logit-alignment}

Intuitively, the raw model $y_i$ outputs for a non-augmented input $x_i$ should produce similar outputs as the saliency-preserving outputs $\tilde{y_i}$ obtained for an augmented sample $\tilde{x_i}$. Given the superfluous, or otherwise non-essential information is removed from the input, the resulting output logits should be similar. Conversely, we expect the $y_i'$ to be similar to $\tilde{y_i}'$ since the salient information is degraded from both samples. Without the latter constraint, models may appear to be aligned with human salient features, but in realty may marshal an entirely different neurons to classify the sample in unexpected ways.

The injection of saliency-based noise (\ie Gaussian blur) may have effects on output logit space, skewering the probability distribution in ways that may obfuscate the metric to assess both distributions. Specifically, the noise may induce changes within the scale parameters (as opposed to simple shifts, otherwise location parameters).
These challenges are consistent with knowledge distillation techniques, which distill knowledge from larger capacity networks into smaller networks by similarly aligning output distributions. Consistent with these works, we use a softmax temperature scaling to increase the effect of lower probabilities, which flattens the probability distribution. Each logit is divided by $\tau$ temperature constant before the softmax is calculated, and the output is multiplied by $2^\tau$. Following previous work \cite{hinton2015distilling}, we set $\tau=2$ unless stated otherwise and hold it constant throughout all experiments.

We penalize the divergence between the aforementioned pairs of logit probability distributions using a Jensen-Shannon (JS) divergence:

\begin{equation}
    {JS(y_s||\tilde{y_s})}=\frac{1}{2}D(y_s||m)+\frac{1}{2}D(\tilde{y_s}||m)
\end{equation}

\noindent where $y_s$ is the model logits of $f(x)$, $\tilde{y_s}$ is the logits of $f(\tilde{x})$, $m$ is a mixture distribution between $y_s$ and $\tilde{y_s}$ defined as $m=\frac{1}{2}(y_s+\tilde{y_s})$, and $D$ is the Kullback–Leibler (KL) divergence.

Conversely, we also match the probability distributions between the fully-degraded input $y_s'$ and the salient-degraded regions of the input $\tilde{y_{s}}'$:

\begin{equation}
    {JS(y_s'||\tilde{y_s}')}=\frac{1}{2}D(y_s'||m')+\frac{1}{2}D(\tilde{y_s}'||m')
\end{equation}

\noindent where $y'_s$ is the model logits of $f(x)$, $\tilde{y}'{_s}$ is the logits of $f(\tilde{x})$, and $m'$ is a mixture distribution between $y'_s$ and $\tilde{y}'{_s}$ defined as $m'=\frac{1}{2}(y'_s+\tilde{y}'{_s})$.

We opt for JS over the more prevalent KL divergence due to its probabilistic symmetry and numerical smoothness \cite{cai2022distances, nielsen2019jensen}. Unlike KL divergence, JS divergence is symmetric as it considers both probability distributions equally ($D(P||Q)$ and $D(Q||P)$, as opposed to simply $D(P||Q)$), which is crucial to our experiments as we are degrading the input, not reducing it to complete noise. Though the high-frequency information is removed from the input, there is still a meaningful, but degraded probability distribution that may be extracted. Second, JS is a numerically smoother compared to KL as it incorporates a mixture of both probability distributions ($M=\frac{1}{2}(P+Q)$). The inclusion of this mixture distribution is what differentiates it from Jeffrey's divergence, which is purely a symmetric KL-divergence \cite{jeffreys1998theory}. Given we our matching probability distributions among several loss surfaces, we find JS to serve as a built-in, soft regularization component within our loss. Preliminary experiments indicated that KL and Jeffrey's divergence performed less favorable than JS, but still effective in increasing performance.

Finally, we blend both components equally into the saliency-based logit alignment (SLA) loss component:

\begin{equation}
\begin{split}
\mathcal{L}_\text{SLA} = (\lambda)(2^{\tau}\cdot JS(y_s||\tilde{y_s})) + \\
(1-\lambda)(2^{\tau}\cdot JS(y_s'||\tilde{y_s}'))
\end{split}
\end{equation}

\noindent where $\lambda$ represents the weighting between salient and non-salient distributions. For simplicity, we weight these equally and set $\lambda=0.5$ for all our experiments. However, we consider $\lambda$ a human-designed prior based upon the evaluation environment and experimental considerations.

\subsection{Saliency-Guided Contrastive Embeddings}
\label{sec:contrastive-embeddings}

We also guide the model towards salient features using a triplet margin contrastive loss \cite{schroff2015facenet}. We do so by using a forward hook at a predefined target layer in the network. For fair comparison across various architectures, we use the embeddings preceding the classification layer, which is often an adaptive pooling layer to flatten the feature maps for the classification head.

We define Salient-Guided Contrastive Embeddings (SCE) as:

\begin{equation}
\begin{split}
\mathcal{L}_\text{Triplet}(z, \tilde{z}, \tilde{z}') = \max\{cs(z,\tilde{z})-cs(z,\tilde{z}')+\mu,0\}
\end{split}
\end{equation}

\noindent where $cs$ is the cosine distance function ($=1-$cosine similarity), $z$ is the anchor, $\tilde{z}$ is the positive, and $\tilde{z}'$ is the negative sample. We found that other similarity measures like angular distance were numerically unstable during training due to the division by $\pi$, which would require a modification of the loss to incorporate more smoothing properties.

Unlike previous works, we use a dynamic, batch-specific margin $m$ between positive and negative samples that represents the cosine distance between $z$ and $z'$.

We also consider the inverse, defined by:

\begin{equation}
\begin{split}
\mathcal{L}_\text{Triplet}(z', \tilde{z}', \tilde{z}) = \max\{cs(z',\tilde{z}') \\
- cs(z',\tilde{z})+\mu,0\}
\end{split}
\end{equation}

Finally, we weight the importance of salient and non-salient guidance and sum them together to form the saliency-guided contrastive embeddings (SCE) loss:
\vskip-4mm
\begin{equation}
\mathcal{L}_\text{SCE} = (\lambda)\mathcal{L}_\text{Triplet}(z, \tilde{z}, \tilde{z}')
+ {(1-\lambda)\mathcal{L}_\text{Triplet}(z', \tilde{z}', \tilde{z})}
\end{equation}
\noindent where $\lambda$ (set in our experiments to 0.5) controls the symmetry between the embeddings guidance and the logit matching probability distribution.

\subsection{The SAGE Loss}
\label{sec:sage-loss}

For completeness, we define the cross-entropy loss component as:
\begin{equation}
\mathcal{L}_\text{Xent} = \sum_{c=1}^{C}\bm{1}_{y \in \mathcal{C}_c} \log p\big(y \in \mathcal{C}_c\big)
\end{equation}

\noindent where $y$ is a class label, $\bm{1}$ is a class indicator function equal to $1$ when $y \in \mathcal{C}_c$ (and 0 otherwise), and $C$ is the total number of classes.

Finally, we define the \text{SAGE} loss as:
\vskip-4mm
\begin{equation}
\label{equation:cosine-final}
\mathcal{L}_\text{SAGE} = (\alpha)\mathcal{L}_\text{Xent} + (1 - \alpha)({\mathcal{L}_\text{SLA} + \mathcal{L}_\text{SCE})}
\end{equation}

\noindent where $\alpha$ (set in our experiments to 0.5) is a tradeoff between the classification loss and the saliency loss components.

\section{Methodology}

\subsection{Datasets}

We provide expansive experiments upon the performance of our method against a variety of datasets across several domains and classification scenarios. We explore open-set classification performance using iris presentation attack detection (Iris PAD), chest X-ray anomaly detection (Chest), and synthetic face detection (Face). These datasets include rich saliency provided by human experts, collected from either mouse annotations or eye-tracking.

\paragraph{Iris Presentation Attack Detection} Iris samples were samples from a larger repository of bona fide and spoof iris samples
\cite{casia-database,Sung_OE_2007,Galbally_ICB_2012,Kohli_ICB_2013,Yambay_ISBA_2017,Trokielewicz_IVC_2020,Kohli_BTAS_2016,Wei_ICPR_2008,Trokielewicz_BTAS_2015,das2020iris}. We evaluate using the Iris Liveness 2020 Competition \cite{das2020iris}, consistent with prior saliency-based works \cite{crum2023mentor, crum2024grains}.

\paragraph{Chest X-ray Anomalies} Normal and abnormal chest X-ray radiographs were sampled from Johnson \etal \cite{johnson2019mimic}. Abnormal radiographs included patients diagnosed with one or several abnormalities: atelectasis (collapse of the lung), cardiomegaly (enlarged heart), pleural effusion (buildup of fluid),  lung opacity (abnormal tissue density), pneumonia, edema (swelling), or support devices such as pacemakers.

\paragraph{Synthetic Face Detection} We follow previous works and include synthetic face detection offered by Boyd \etal \cite{boyd2021cyborg}. Real faces are sampled from FFHQ and Celeb-HQ, whereas synthetically generated faces are provided by an assortment of StyleGAN-based generators from previous work \cite{boyd2021cyborg}.

\subsection{Experimental Design}

\paragraph{Baselines \& Training Configurations} We compare our method across a wide variety of saliency-based methods. Loss-only guidance includes CYBORG \cite{boyd2021cyborg}, Unet+Gaze \cite{karargyris2021creation}, GG-CAM \cite{zhu2022gaze}. MENTOR \cite{crum2023mentor} is a pretraining method which guides network embeddings throughout the entire architecture using an two-stage training strategy. We also include  Gaze-GNN \cite{wang2024gazegnn}, a graph neural network that incorporates saliency using embeddings from multiple encoders. Finally, we benchmark against the saliency-guided augmentations introduced in Boyd \etal \cite{boyd2022human}, which includes blurring the non-salient regions of the input samples to guide models towards salient features.

For a fair comparison, each method is trained using identical configurations following previous works \cite{boyd2021cyborg,crum2024grains,crum2023mentor}. Models are initialized with ImageNet weights and trained with a batch size of $20$ for $50$ epochs using stochastic gradient descent. We set the learning rate to $0.005$ and use a  weight decay of $1\times10^{-6}$. Every 12 epochs a scheduler reduces the learning rate by a factor of $0.1$.

\paragraph{Evaluation Measures} For binary classification tasks, we use Area Under the Receiver Operating Characteristic Curve (AUROC) following previous works.

%% file: figures/Face-Sampler.tex
\begin{figure*}[t]
  \centering
  \includegraphics[width=\linewidth]{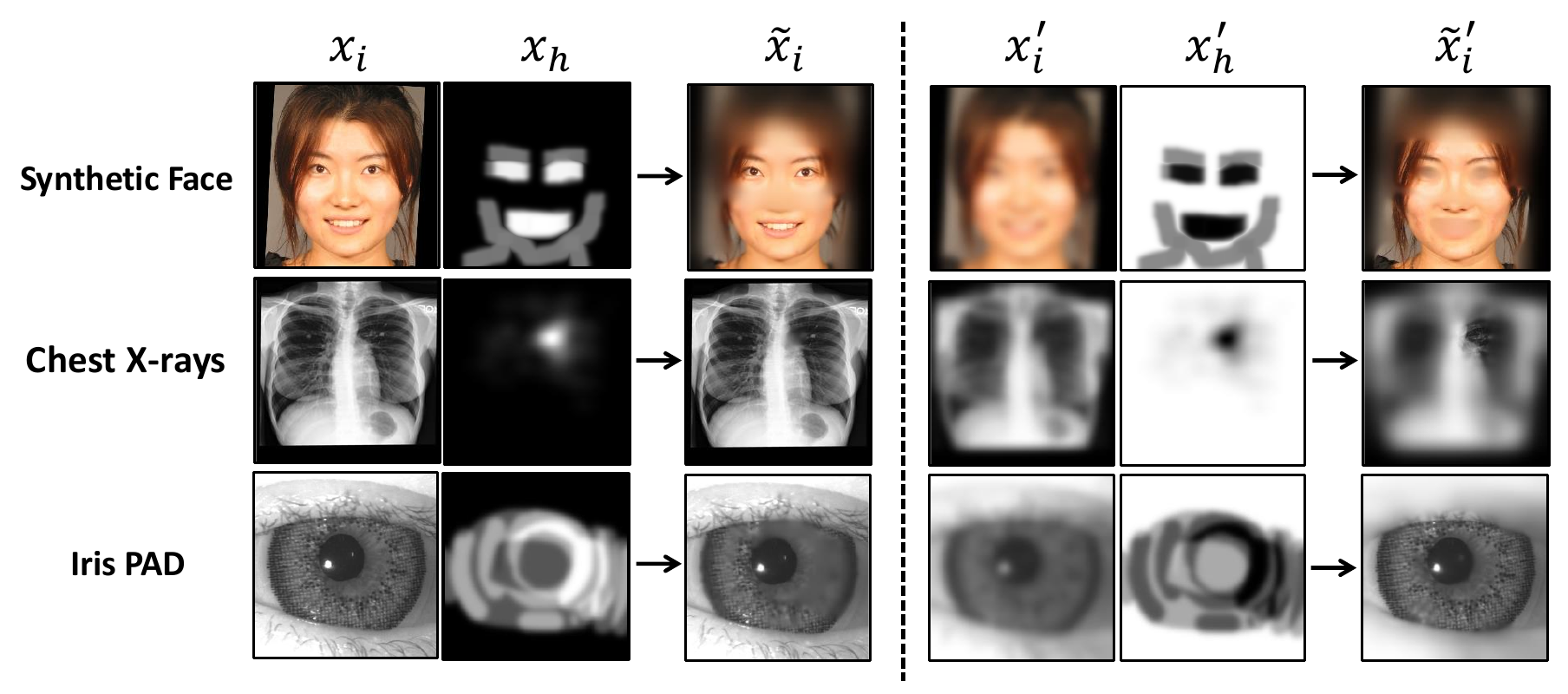}
  \caption{Examples and notation of the saliency-based augmentations described in this paper (left to right): input sample $x_i$, human saliency map (indicating salient regions) $x_h$, selective blurring that preserves the salient regions $\tilde{x_i}$. Right side: blurring on entire image $x'_i$, the inverse of saliency map (indicating non-salient regions) $x'_h$, selective blurring that degrades the salient regions $\tilde{x_i}'$.}
  \label{fig:sage-notation-dataset}
  \vspace{-5mm}
\end{figure*}

%% file: latex/04-Results.tex
\input{tables/open-set}

\section{Results}
\label{sec:sage-results}

This section first begins the generalization performance of our method, including the addition of different architectures, such as vision transformers. Next, we include a set of sanity check experiments to ensure the proposed method is working as expected. Finally, we conclude with an ablation study highlighting the affect temperature scaling has on performance.

\subsection{Generalization Performance}
SAGE increases the open-set generalization performance compared to existing SOTA saliency-based methods for all three datasets for the ResNet50 architecture (see Tab.\ref{tab:sage-open-set-results}). SPecifcially, SAGE increases performance of iris-PAD (0.945), with MENTOR being the closest at 0.909. 
The remaining saliency-based methods either perform marginally better than the Baseline at 0.878 (CYBORG 0.886), or even worse (Unet+Gaze 0.861, GG-CAM 0.615, Gaze-GNN 0.842). Saliency-guidance through embeddings is substantial compared to existing methods, followed by pretraining, loss-based, and architectural. Moreover, SAGE achieves as similar comparative performance for chest X-ray anomaly detection (0.850), holding a modest gain over MENTOR (0.844), CYBORG (0.840) and the Baseline (0.832).

\input{tables/backbones}

Interestingly, SAGE reports a more modest performance for the synthetic face detection task (0.587), followed by CYBORG (0.531), Gaze-GNN (0.524), and MENTOR (0.506). These results suggest that SAGE may not be as effective on datasets that already include high-frequency information associated with synthetic signatures, which is not appropriately captured by the saliency provided to the model and consistent with previous findings \cite{boyd2021cyborg, crum2023mentor}. Given the non-salient portions of the input may be embedded with generator-dependent signals, thereby undermining the effectiveness of the augmentation strategy and its ability to guide the network's embeddings.

\paragraph{Network Architectures} We extend our analysis to include different neural network architectures to validate our initial findings (see Tab.\ref{tab:sage-ablation-backbones}). SAGE raises generalization for EfficientNet-b7 backbones for Iris and Chest X-rays over existing methods \cite{tan2019efficientnet}, and DenseNet-121 for Iris \cite{huang2017densely}. Our experiments indicate that our method broadly generalizes to vision transformers, which tend to struggle in performance compared to CNNs on these datasets due to the limited supply of training data. Nevertheless, SAGE is able to increase the capacity of Swin transformer for two of the three datasets \cite{liu2021swin}. However, Swin struggled to converge for synthetic face detection, likely requiring more data or different training configurations before more analysis can be performed.

\subsection{Sanity Checks with Flipped Saliency Maps}
\input{tables/sanity-check-results}

An often overlooked experiment within saliency-based methods is performing rudimentary sanity checks within the saliency maps. Intuitively, if the model's improvement in performance is due to the guidance towards salient features, the performance should decrease when the saliency maps are reversed (\ie saliency maps indicate non-salient regions of the input). SAGE has the highest drop in performance with inverted saliency maps, which provides empirical evidence that it works by guidance towards salient features, not by pure regularization. SAGE recorded the largest difference in performance among all methods compared for Iris PAD (0.945$\rightarrow$0.906), followed by Chest X-ray anomalies (0.850$\rightarrow$0.836) and Synthetic Face Detection (0.587$\rightarrow$0.577).

We hypothesize that the reasoning behind these results is that both Unet+Gaze and MENTOR are explicitly disconnected from the classification layer and it has no mechanism to connect both loss components. Unet+Gaze's classifier uses the embeddings after the encoder, which neglects the reconstruction of salient features by the decoder. Additionally, MENTOR's pretraining only guides the network where to look, not what those regions mean (\ie classification of those regions). Conversely, CYBORG's use of the predicted class CAM provides the network with some connection between the classifier and the saliency maps, but yet offers guidance when the classifier generates incorrect CAMs. SAGE further improves this relationship by imposing more structure through its contrastive loss, which provides encouragement towards salient features and discouragement towards the non-salient. Moreover, the use of saliency-guided augmentations and logits alignment ensures the network is incorporating the human saliency maps into its classifications, from the beginning of the model (input) to the very end (logits).

\subsection{Ablation Study}
\input{tables/ablation-study}

We also include a small set of experiments exploring the role temperature scaling ($\tau$) may have on generalization performance (see Tab. \ref{tab:sage-ablation-study}). Temperature is a scaling mechanism to determine the uniformity of our logit distributions. Higher values indicate a flatter probability distribution, which relaxes the constraint on the network and encourages greater exploration of the solution space. When $\tau=1$, the distribution remains unchanged. Our experiments suggest that $\tau=2$ is optimal for both Iris and Face, whereas Chest had a slight bump in performance for a higher temperature. Additionally, it may also be an indication of the quality of information provided by the saliency maps. For example, higher temperature values may be beneficial when saliency is provided by experts (\eg radiologists for chest X-ray anomalies), and lower values for non-experts or less experienced annotators. Interestingly, this may be a way to change the granularity of the input sample beyond the normalized image space. Our performance is on par with previous work in studying the granularity of saliency collection methods \cite{crum2024grains}, which may be a fascinating connection within the hyperparameter of network training as opposed to the input space. Though not a core part of our work, we leave interesting questions related to the annotation quality for future work.

Finally, we include a small set of experiments exploring varied $\sigma$ levels (see Tab. \ref{tab:sage-ablation-study-sigma}). For Iris PAD, performance increased for higher $\sigma$ amounts, whereas chest X-rays tend to benefit from more moderate amounts. Finally, synthetic face detection was optimal when $\sigma=10$ and performing mildly given higher or lower amounts.

\input{tables/ablation-study-sigma}

%% file: tables/open-set.tex
\begin{table}[h]
\scriptsize
\centering
\caption{Open-set datasets with \textbf{ResNet50} backbone. Means and standard deviations of Area Under the ROC Curve (AUC) is reported across 5 independent train-test runs.}
\label{tab:sage-open-set-results}
\begin{tabular}{@{}l|lll@{}}
\toprule
\textbf{Method} & \textbf{Iris} & \textbf{Face} & \textbf{Chest} \\
\midrule
Baseline (Xent) & 0.878\tiny{$\pm$0.018} & 0.487\tiny{$\pm$0.064} & 0.832\tiny{$\pm$0.009}  \\
CYBORG \cite{boyd2021cyborg} & 0.886\tiny{$\pm$0.020} & 0.531\tiny{$\pm$0.050} & 0.840\tiny{$\pm$0.019} \\
UNET+Gaze \cite{karargyris2021creation} & 0.861\tiny{$\pm$0.016} & 0.409\tiny{$\pm$0.048} & 0.835\tiny{$\pm$0.013} \\
Gaze-GNN \cite{wang2024gazegnn} & 0.842\tiny{$\pm$0.027} & 0.524\tiny{$\pm$0.046} & 0.796\tiny{$\pm$0.013} \\
MENTOR \cite{crum2023mentor} & 0.909\tiny{$\pm$0.023} & 0.506\tiny{$\pm$0.044} & 0.844\tiny{$\pm$0.006} \\
GG-CAM \cite{zhu2022gaze} & 0.615\tiny{$\pm$0.052} & 0.452\tiny{$\pm$0.043} & 0.784\tiny{$\pm$0.002} \\
Boyd \etal \cite{boyd2022human} & 0.834\tiny{$\pm$0.021} & 0.464\tiny{$\pm$0.080} & 0.803\tiny{$\pm$0.011} \\
\midrule
\textbf{SAGE (ours)} & \textbf{0.945\tiny{$\pm$0.012}} & \textbf{0.587\tiny{$\pm$0.060}} & \textbf{0.850\tiny{$\pm$0.011}} \\
\bottomrule
\end{tabular}
\end{table}

%% file: tables/backbones.tex
\begin{table*}[t]
\scriptsize
\centering
\caption{Open-set datasets with various backbones. Means and standard deviations of \textbf{AUC} is reported across 5 independent runs.}
\label{tab:sage-ablation-backbones}
\begin{tabular}{@{}l|l|ccc@{}}
\toprule
\textbf{Method} & \textbf{Backbone} & \textbf{Iris} & \textbf{Face} & \textbf{Chest} \\
\midrule
Baseline (Xent) & ResNet50 & 0.878\tiny{$\pm$0.018} & 0.474\tiny{$\pm$0.080} & 0.832\tiny{$\pm$0.009}  \\
 & DenseNet121 & 0.881\tiny{$\pm$0.016} & 0.508\tiny{$\pm$0.042} & 0.829\tiny{$\pm$0.015} \\
 & EfficientNet-b7 & 0.883\tiny{$\pm$0.012} & 0.465\tiny{$\pm$0.020} & 0.817\tiny{$\pm$0.010} \\

 & Swin-T & 0.678\tiny{$\pm$0.036} & 0.472\tiny{$\pm$0.018} & 0.810\tiny{$\pm$0.024} \\
\midrule
CYBORG \cite{boyd2021cyborg} & ResNet50 & \text{0.886\tiny{$\pm$0.020}} & 0.531\tiny{$\pm$0.050} & 0.840\tiny{$\pm$0.019} \\
 & DenseNet-121 & 0.895\tiny{$\pm$0.018} & \textbf{0.617\tiny{$\pm$0.027}} & \textbf{0.852\tiny{$\pm$0.003}} \\
 & EfficientNet-b7 & 0.800\tiny{$\pm$0.025} & 0.500\tiny{$\pm$0.030} & 0.820\tiny{$\pm$0.010} \\
\midrule
Unet+Gaze \cite{karargyris2021creation} & ResNet50 & 0.861\tiny{$\pm$0.019} & 0.490\tiny{$\pm$0.054} & 0.835\tiny{$\pm$0.013} \\
 & DenseNet-121 & 0.860\tiny{$\pm$0.013} & 0.505\tiny{$\pm$0.030} & 0.828\tiny{$\pm$0.006} \\
 & EfficientNet-b7 & 0.830\tiny{$\pm$0.039} & 0.453\tiny{$\pm$0.086} & 0.829\tiny{$\pm$0.009}\\
 & Swin-T & 0.484\tiny{$\pm$0.018} & 0.463\tiny{$\pm$0.007} & 0.681\tiny{$\pm$0.069} \\
\midrule
MENTOR \cite{crum2023mentor} & ResNet50 & 0.909\tiny{$\pm$0.023} & 0.528\tiny{$\pm$0.061} & 0.844\tiny{$\pm$0.006} \\
 & DenseNet-121 & 0.893\tiny{$\pm$0.008} & 0.565\tiny{$\pm$0.035} & \text{0.841\tiny{$\pm$0.009}} \\
  & EfficientNet-b7 & 0.895\tiny{$\pm$0.021} & \textbf{0.565\tiny{$\pm$0.030}} & 0.837\tiny{$\pm$0.005}\\
 & Swin-T & 0.698\tiny{$\pm$0.079} & 0.439\tiny{$\pm$0.014} & 0.834\tiny{$\pm$0.013} \\
\midrule
Boyd \etal \cite{boyd2022human} & ResNet50 &  0.834\tiny{$\pm$0.021} & 0.464\tiny{$\pm$0.080} & 0.803\tiny{$\pm$0.011} \\
& DenseNet-121 & 0.845\tiny{$\pm$0.016} & 0.460\tiny{$\pm$0.042} & 0.817\tiny{$\pm$0.015} \\
& EfficientNet-b7 & 0.813\tiny{$\pm$0.019} & 0.455\tiny{$\pm$0.025} & 0.670\tiny{$\pm$0.040} \\
& Swin-T &  0.697\tiny{$\pm$0.013} & 0.432\tiny{$\pm$0.021} & 0.798\tiny{$\pm$0.018} \\
\midrule
SAGE (ours) & ResNet50 & \textbf{0.945\tiny{$\pm$0.012}} & \textbf{0.587\tiny{$\pm$0.060}} & \textbf{0.850\tiny{$\pm$0.011}} \\
 & DenseNet-121 & \textbf{0.920\tiny{$\pm$0.015}} & \text{0.584\tiny{$\pm$0.035}} & \text{0.847\tiny{$\pm$0.006}} \\
  & EfficientNet-b7 & \textbf{0.913\tiny{$\pm$0.008}} & 0.468\text{\tiny{$\pm$0.016}} & \textbf{0.841\textbf{\tiny{$\pm$0.004}}} \\
 & Swin-T & \textbf{0.722\tiny{$\pm$0.018}} & 0.422\tiny{$\pm$0.017} & \textbf{0.847\tiny{$\pm$0.008}} \\
\bottomrule
\end{tabular}
\vspace{-3mm}
\end{table*}

%% file: tables/sanity-check-results.tex
\begin{table*}[t]
\scriptsize
\centering
\caption{\textbf{Sanity check} for each method using the \textbf{ResNet50} backbone. \textbf{Saliency Maps} indicate the original saliency maps ($x_h$), whereas \textbf{Non-Salient Maps} are reversed so that the non-salient features are annotated as opposed to the salient-features ($x'_h$). For an overview of the notation, please see Fig. \ref{fig:sage-notation-dataset}. Arrows indicate whether higher or lower values are better. Means and standard deviations of \textbf{AUC} is reported across 5 independent runs.}
\label{tab:sage-sanity-check}
\begin{tabular}{@{}c|c|c|c|c@{}}
\toprule
\multirow{2}{*}{\bf Method} &\multirow{2}{*}{\bf Dataset} & \bm{$x_h$} & \bm{$x'_h$} & \bm{$\Delta$} \\
 & \textbf{} & \textbf{Saliency Maps} \bm{$\uparrow$} & \textbf{Non-Salient Maps} \bm{$\uparrow$} & \textbf{Difference}\bm{$\uparrow$}  \\
\midrule
CYBORG \cite{boyd2021cyborg} & Iris & 0.886\tiny{$\pm$0.020} & 0.888\tiny{$\pm$0.011} & -0.002 \\
 & Face & 0.531\tiny{$\pm$0.050} & 0.525\tiny{$\pm$0.031} & 0.006 \\
 & Chest & 0.840\tiny{$\pm$0.019} & 0.836\tiny{$\pm$0.007} & 0.004 \\
 \midrule
Unet+Gaze \cite{karargyris2021creation} & Iris & 0.861\tiny{$\pm$0.019} & 0.882\tiny{$\pm$0.022} & -0.021 \\
 & Face & 0.490\tiny{$\pm$0.054} & 0.508\tiny{$\pm$0.048} & -0.018 \\
 & Chest & 0.835\tiny{$\pm$0.013} & 0.842\tiny{$\pm$0.005} & -0.007 \\
\midrule
MENTOR \cite{boyd2021cyborg} & Iris & 0.909\tiny{$\pm$0.023} & 0.893\tiny{$\pm$0.010} & 0.016 \\
 & Face & 0.528\tiny{$\pm$0.061} & 0.458\tiny{$\pm$0.052} & \textbf{0.07} \\
 & Chest & 0.844\tiny{$\pm$0.006} & 0.842\tiny{$\pm$0.005} & 0.002 \\
\midrule
\text{SAGE (ours)} & Iris & \text{0.945\tiny{$\pm$0.012}} & 0.906\tiny{$\pm$0.013} & \textbf{0.039} \\
 & Face & \text{0.587\tiny{$\pm$0.060}} & 0.577\tiny{$\pm$0.037} & \text{0.010} \\
 & Chest & \text{0.850\tiny{$\pm$0.011}} & 0.836\tiny{$\pm$0.004} & \textbf{0.014} \\
\bottomrule
\end{tabular}
\vspace{-2mm}
\end{table*}

%% file: tables/ablation-study.tex
\begin{table}[t]
\scriptsize
\centering
\caption{Ablation study of \bm{$\tau$} and probability distribution for \textbf{ResNet50} backbone. Means and standard deviations of \textbf{AUC} is reported across 5 independent runs.}
\label{tab:sage-ablation-study}
\begin{tabular}{@{}l|lll@{}}
\toprule
\bm{$\tau$} & \textbf{Iris} & \textbf{Face} & \textbf{Chest} \\
\midrule
 1 & 0.935\tiny{$\pm$0.009} & \text{0.573\tiny{$\pm$0.032}} & 0.837\tiny{$\pm$0.011} \\
  2 & \textbf{0.945\tiny{$\pm$0.012}} & \textbf{0.587\tiny{$\pm$0.060}} & 0.850\tiny{$\pm$0.011} \\ 
  3 & 0.940\tiny{$\pm$0.010} & 0.538\tiny{$\pm$0.020} & \textbf{0.852\tiny{$\pm$0.003}} \\
\bottomrule
\end{tabular}
\vspace{-5mm}
\end{table}

%% file: tables/ablation-study-sigma.tex
\begin{table}[t]
\scriptsize
\centering
\caption{Ablation study of \bm{$\sigma$} and probability distribution for \textbf{ResNet50} backbone. Means and standard deviations of \textbf{AUC} is reported across 5 independent runs.}
\label{tab:sage-ablation-study-sigma}
\begin{tabular}{@{}l|l|lll@{}}
\toprule
\bm{$\sigma$} & \textbf{Iris} & \textbf{Face} & \textbf{Chest} \\
\midrule
 5 & 0.929\tiny{$\pm$0.008} & 0.551\tiny{$\pm$0.058} & 0.847\tiny{$\pm$0.002} \\
 7 & 0.935\tiny{$\pm$0.007} & 0.546\tiny{$\pm$0.021} & \textbf{0.851\tiny{$\pm$0.009}} \\
 10 & \textbf{0.945\tiny{$\pm$0.012}} & \textbf{0.587\tiny{$\pm$0.060}} & 0.850\tiny{$\pm$0.011} \\
 12 & 0.944\tiny{$\pm$0.010} & 0.545\tiny{$\pm$0.039} & 0.844\tiny{$\pm$0.010} \\
\bottomrule
\end{tabular}
\end{table}

%% file: latex/05-Conclusion.tex
\section{Conclusions}
\label{sec:sage-conclusion}

The incorporation of human salient information into deep learning is nontrivial and often replete with challenges related to heatmap normalization, architecture-specific, and unreliable model-generated saliency maps. Our insight is that these challenges stem in part from the $L_2$ norm. This paper presents SAGE, an approach that avoids these challenges by operating exclusively within the model's own embedding space. We propose a loss function that incorporates how the model is affected by salient-based augmentations, which is used as a steering mechanism within a contrastive triplet loss. In addition to model guidance, we enforce the model's respective outputs to match human intuition as observed within the natural space. Our method is agnostic to any neural network encoding architecture and model saliency method, paving the way for use on more widespread tasks and domains.